\documentclass{article}
\usepackage{spconf,amsmath,graphicx}
\usepackage{bbding}
\usepackage{float}
\usepackage{balance}
\usepackage{cite}
\usepackage{caption}
\title{Towards Reliable Image Outpainting: Learning Structure-Aware Multimodal Fusion with Depth Guidance}
\name{Lei Zhang \qquad Chunyu Lin \qquad Kang Liao \qquad Yao Zhao}

\address{The Institute of Information Science, Beijing Jiaotong University, Beijing 100044, China}
\begin{document}
%
\maketitle
\begin{abstract}
Image outpainting technology generates visually plausible content regardless of authenticity, making it unreliable to be applied in practice. Thus, we propose a reliable image outpainting task, introducing the sparse depth from LiDARs to extrapolate authentic RGB scenes. 
The large field view of LiDARs allows it to serve for data enhancement and further multimodal tasks.
Concretely, we propose a Depth-Guided Outpainting Network to model different feature representations of two modalities and learn the structure-aware cross-modal fusion.
And two components are designed: 1) The \textit{Multimodal Learning Module} produces unique depth and RGB feature representations from the perspectives of different modal characteristics. 2) The \textit{Depth Guidance Fusion Module} leverages the complete depth modality to guide the establishment of RGB contents by progressive multimodal feature fusion.
Furthermore, we specially design an additional constraint strategy consisting of \textit{Cross-modal Loss} and \textit{Edge Loss} to enhance ambiguous contours and expedite reliable content generation. 
Extensive experiments on KITTI and Waymo datasets demonstrate our superiority over the state-of-the-art method, quantitatively and qualitatively.  
\end{abstract}
\section{Introduction}
\label{sec:intro}

\begin{figure}
  \centering
  \includegraphics[width=1.0\linewidth]{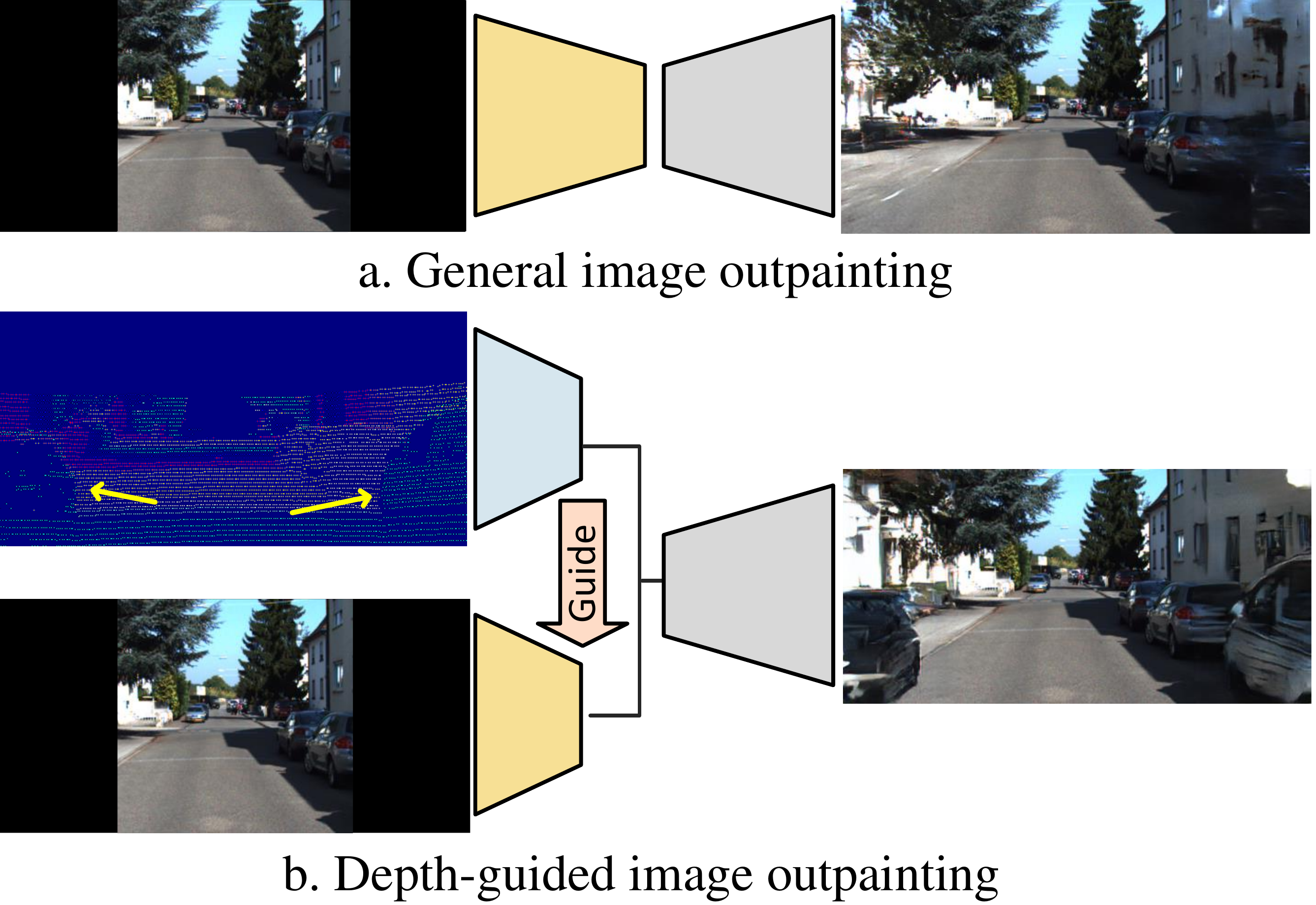}
  \setlength{\abovecaptionskip}{-0.2cm}  
  \setlength{\belowcaptionskip}{-0.5cm} 
  \caption{A comparison of image outpainting tasks: (a) general image outpainting and (b) the proposed depth-guided image outpainting.}
  \label{teaser}
\end{figure}

Image outpainting aims to extrapolate the unknown image area, generating reasonable content based on the original style and structure. 
But the general outpainting is fictitious and theoretical like Fig.\ref{teaser}(a).
In recent studies, researchers\cite{ kim2021painting, bowen2021oconet, khurana2021semie, wang2021sketch, zhang2021image} propose to introduce additional modal information to the general image outpainting. 
They focus on blending various modal contents to reach better outpainting performance for respective target scenes. For example, Zhang\cite{zhang2021image} $et$ $al.$ introduced post-processed dense depth to image outpainting. However, they ignore the differences between modalities and their inputs have to be generated manually by the stereo camera and LiDARs.
To address the above issue, we propose to additional employ only LiDAR data. It reflects the distribution of scenes naturally and has a broader field of view than the camera. 
And we obtain sparse depth maps by projection to achieve the reliable image outpainting task like Fig.\ref{teaser}(b). 
Besides, the task shows new vigor and potential vitality for outpainting to serve data enhancement and practical multimodal applications\cite{hwang2021LiDAR}. 

To this end, we propose the Depth-Guided Outpainting Network (DGONet) to learn the structure-aware multimodal fusion with sparse depth.
The sparse depth and RGB have two significant modal differences: 1) the pixel values carry different messages --- depth and intensity; 2) the valid pixels in the depth map from LiDARs are significantly sparser than that in the RGB image.
Thus, we design a \textit{Multimodal Learning Module}(ML) to extract unique feature representations from different modalities, contributing to more beneficial and discriminative semantic representations from their low-level features. 
When it comes to the multimodal feature fusion, previous works~\cite{hu2021penet, zhang2021image} roughly adopt the concatenation operation in channels to implement the interaction between these two modal features. They omit the modal roles in different tasks, and the interaction is somewhat weak.
\begin{figure*}[htbp]
\centering
\includegraphics[width=0.8\textwidth]{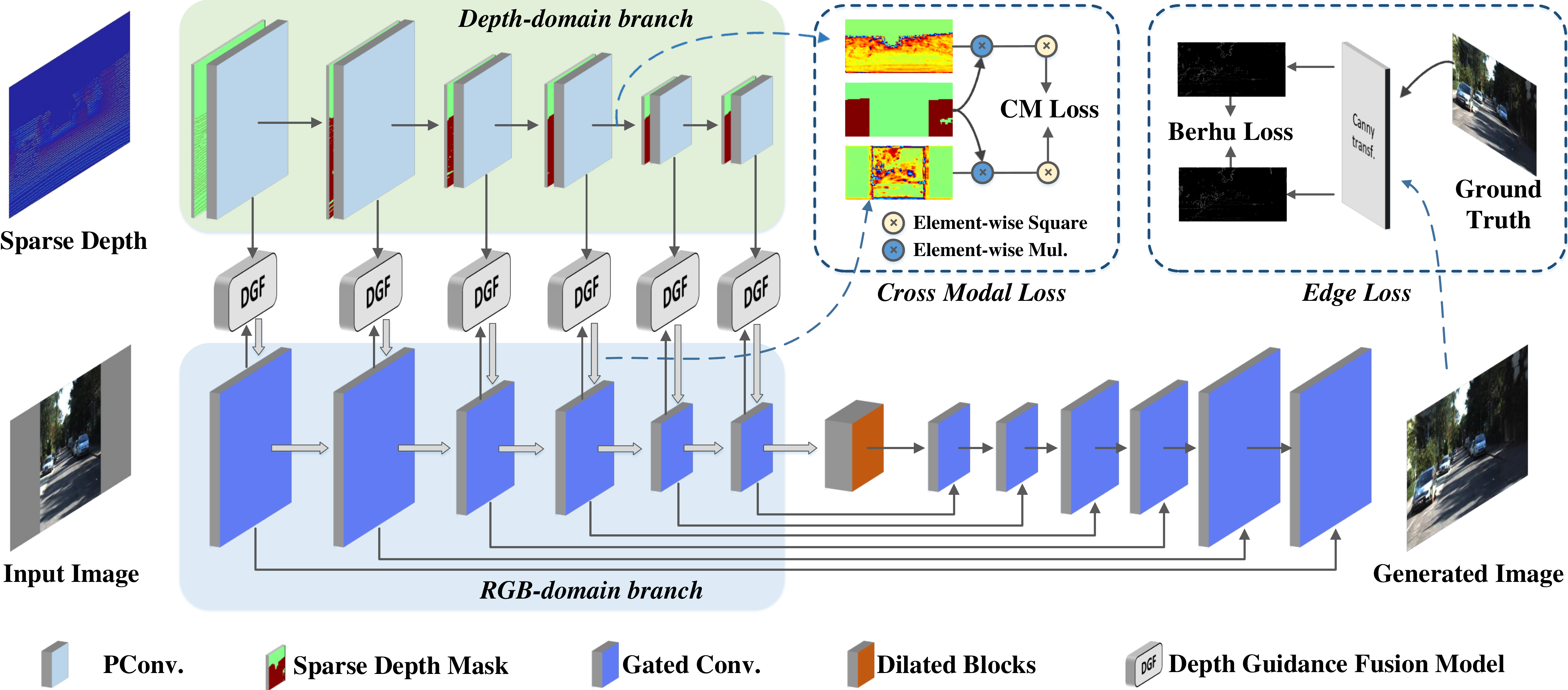}
\centering
  \setlength{\abovecaptionskip}{0.2cm}     \setlength{\belowcaptionskip}{-0.4cm}   
\caption{An overview of the network architecture.}
\label{overview}
\end{figure*}
Instead, we design a \textit{Depth Guidance Fusion Module}(DGF) to encourage the depth modality to facilitate the establishment process of extrapolated RGB contents like Fig.\ref{teaser}(b). 
Specifically, this module adopts a progressive fusion scheme to enhance feature interaction by converting the depth features into dynamic kernels to filter the RGB features, generating semantic-consistent fused features with the depth features.
Considering the significant content deficiency in RGB images, 
we propose a \textit{Cross-modal Loss} to force the extrapolated RGB feature to be similar to the high-level semantics of the depth modality. 
To avoid the boundary-blurring phenomenon occurring continually in generating tasks, we also introduce an \textit{Edge Loss} to constrain the model to generate sharp object outlines in the results.
In addition to the standard metrics for image generation, i.e., PSNR, LPIPS, and FID, we employ 2D vehicle detection(AP) to measure the network's performance on reliability to reflect the consistency with the real surroundings. And 3D object detection(AP3D(R40)) is also used to validate the importance of our network in multimodal applications.
\vspace{-5pt}
\section{Method}
\vspace{-5pt}
\subsection{Multimodal Learning Module}
\vspace{-5pt}
\subsubsection{Depth-domain Branch} 
\begin{figure*}[htbp]
\centering
\includegraphics[width=\textwidth]{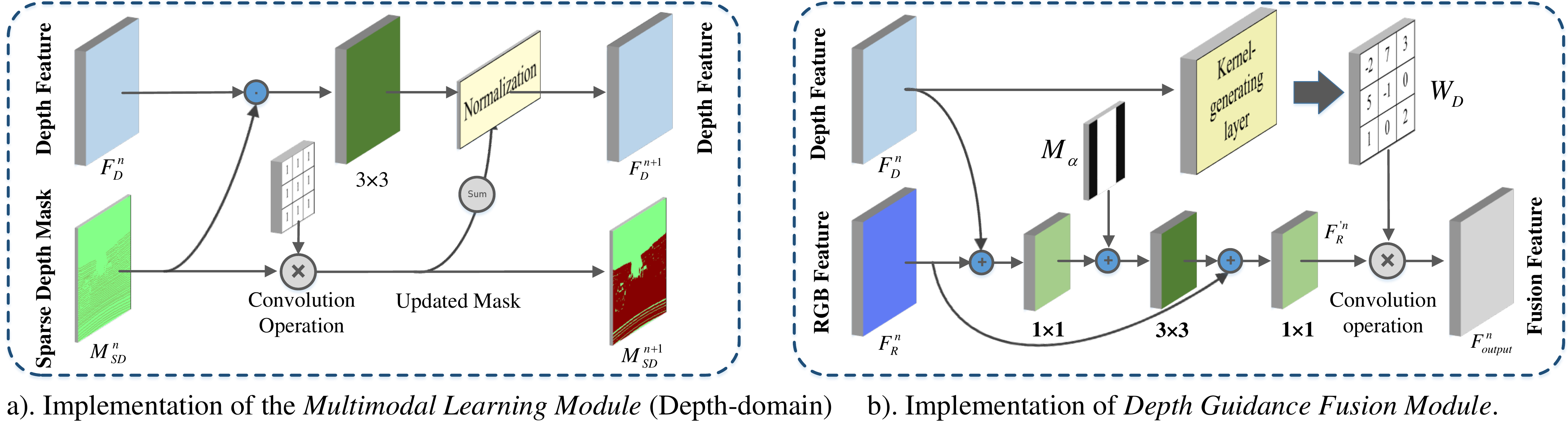}
\centering
  \setlength{\abovecaptionskip}{-0.2cm} 
  \setlength{\belowcaptionskip}{-0.3cm}
\caption{Implementation structure of the main modules.}
\label{modules}
\end{figure*}

\vspace{-5pt}
We take a single channel sparse depth map as Depth-domain input.
But it has an obvious disadvantage that the depth from LiDARs have only {\bfseries 7$\%$ valid pixels}\cite{Uhrig2017THREEDV} on data distribution.
The general convolution treats all pixels equally, so the feature from sparse depth is disrupted by a lot of invalid pixels severely.
To address this, we create a sparse depth mask ${M_{SD}}$ to attenuate the effect of invalid pixels,
and propose partial convolution\cite{liu2018image} in sparse depth like Fig.\ref{modules}(a).
The implementation is formulated as follows:
\begin{equation}
F_D^{{\rm{n + 1}}} = Conv(F_D^{\rm{n}} \otimes M_{SD}^n)\frac{{sum(I)}}{{sum(M_{SD}^{n + 1})}}\end{equation}
where $\otimes$ denotes the element-wise multiplication.
$F_D^{\rm{n}}$ represents the feature of the $n$-th layer in the Depth-domain branch.
$I$ is an all-one matrix with the same shape as ${M_{SD}}$. 
\begin{figure}[ht]
\centering
\includegraphics[width=1\linewidth]{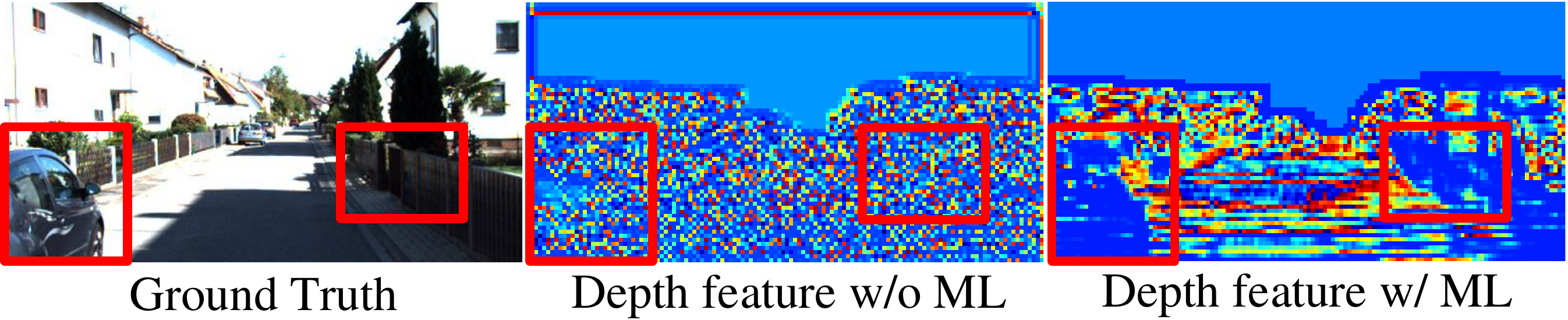}
  \setlength{\abovecaptionskip}{-0.2cm}  
\caption{Feature visualization of the Depth-domain Branch.}
\setlength{\belowcaptionskip}{-1cm} 
\label{fig:ML}
\end{figure}
We count the proportion of valid pixels in the ${M_{SD}}$ as the scaling factor, which is applied as an appropriate normalization for features, adapting to the varying amount of valid inputs and reducing the impact of sparsity. 
Besides, the ${M_{SD}}$ is also updated by the convolution operation along with the encoding process. 
The above encoder design leads to a more significant characterization of the sparse depth structure, specifically distinguishing it from the vanilla convolution in other works, as shown in Fig.\ref{fig:ML}. 
It also provides a more accurate guidance for modal fusion and subsequent decoders.

\vspace{-10pt}
\subsubsection{RGB-domain Branch} 
\vspace{-5pt}
We also set up a binary mask ${M_\alpha }$ of valid RGB pixels and concatenate ${M_\alpha}$ with the three-channel RGB image as inputs. 
As in image generation, the gated convolution\cite{yu2019free} is employed to form the RGB-domain branch, which enables the network to learn a dynamic feature selection mechanism for each channel and spatial location for better filling holes. 

To maintain consistency with the two branches, we set the identical feature size and number of channels in the corresponding layer. 
We merge the RGB features into the decoder by skip connections\cite{ronneberger2015u}, thereby improving the ability of the model to synthesize high-frequency information.


\vspace{-6pt}
\subsection{Depth Guidance Fusion Module}
\vspace{-5pt}
\label{s32}

We reconsider the role of depth for images to design the \textit{Depth Guidance Fusion Module} to restore the authentic RGB context by the depth guidance different from concatenation\cite{zhang2021image, hu2021penet}. 

We take the depth feature $ {F_D^{{\rm{n}}}}$ and the RGB feature $ {F_R^{{\rm{n}}}}$ of the $n$-th layer as inputs. It performs a progressive fusion of the two modalities as shown in Fig.\ref{modules}(b).
We employ ${M_\alpha }$ as an additional channel of the feature to emphasize the network focus on the content generation in the blank mask positions. 
And the basic interaction feature ${{F'_R}^{\rm{n}}}$ is implemented through residuals and channel dimensionality reduction.
Further, inspired by the guided image filtering\cite{he2010guided}, we propose to generate the filtering kernels from intact depth for the image.
In the implementation, we learn directly the kernel weight ${W_D}$ from ${F_D^{{\rm{n}}}}$ by Kernel-Generating Layer\cite{tang2020learning}(implemented by convolution).
The kernel performs further convolution operations with features ${{F'_R}^{\rm{n}}}$ as follows:
\vspace{-3pt}
\begin{equation}
{{F_{output}}^{\rm{n}}} = Conv({{F'_R}^{{\rm{n}}}};{W_D}({F_D};\Theta )) \label{tag}
\end{equation}
where $\Theta $ is the parameter of the Kernel-Generating Layer. 

By the implementations, the kernels are dynamically generated depending on well-structured depth modality .
Therefore, we could employ the depth kernel in the unknown area to build the structure-aware RGB content at the corresponding spatial position and scene. 
These advantages are suitable for our reliable image outpainting task.
\vspace{-5pt}
\subsection{Training Loss}
\subsubsection{Cross-modal Loss}
\vspace{-5pt}
Because the RGB input is incomplete, the learned features by the encoder are barren within the extrapolation region.
While we expect to establish valid content as quickly as possible before the decoder. It is in analogy to the manual painting process requiring enough pigment. 
In consequence, we design the \textit{Cross-modal Loss} between the modal domains as formulated:
\begin{equation}
\phi (F) = \frac{1}{M}\sum\nolimits_i {({F^i} \odot } {F^i})\label{Lcm1}
\end{equation}
\begin{displaymath}
{{\mathcal{L}}_{cm}} = {\left\| {\phi ({F_D} \odot (1 - {{M_\alpha }}) - \phi ({F_R} \odot {M_{SD}} \odot (1 - {{M_\alpha }}))} \right\|_{l2}}
\label{cmloss}
\end{displaymath}
where ${{F^i}}$ represents the $i$-th channel of the feature and $M$ is the number of channels. 
We obtain the attention map by empirically employing sum of squared values referring to the attention mechanism\cite{zagoruyko2016paying}.
Through the constraint, the intact depth modality transfers proper features to the RGB modality. 
It assists the representation of unknown regions more robust and provides more pigments for subsequent layers.
\vspace{-5pt}
\subsubsection{Edge Loss}
\vspace{-5pt}
We also design the constraint on the outlines of the object to eliminate the blurring of generated results.
We feed the output and ground truth into the Canny\cite{canny1986computational} filter to generate the edge maps. 
In the implementation, the edge map is constrained by Berhu loss\cite{zwald2012berhu} to calculate.

\vspace{-5pt}
\subsubsection{Discriminative Loss}
\vspace{-5pt}
Our DGONet includes a discriminator for the adversarial learning.
We likewise introduce conditional vectors\cite{szegedy2016rethinking, teterwak2019boundless} in the discriminator to emphasize that outputs semantically match the ground truth.
And we employ a Wasserstein GAN hingeIn adversarial loss\cite{miyato2018spectral}.
In addition to the loss strategy above, we also introduce a pixel loss to optimize for the coarse image agreement, and is implemented as an $L1$ loss.
The total loss of our framework can be described as:
\begin{equation} 
{{\mathcal{L}}_{total}} = {\lambda _{adv}}{{\mathcal{L}}_{adv,G}} + {\lambda _p}{{\mathcal{L}}_{pixel}} + {\lambda _e}{{\mathcal{L}}_{edge}} + {\lambda _{cm}}{{\mathcal{L}}_{cm}}
\end{equation}
where ${\lambda _{adv}}$, ${\lambda _{edge}}$ and ${\lambda _{cm}}$ are the trade-off factor weights.
\begin{figure*}[htbp]
  \centering
  \includegraphics[width=\textwidth]{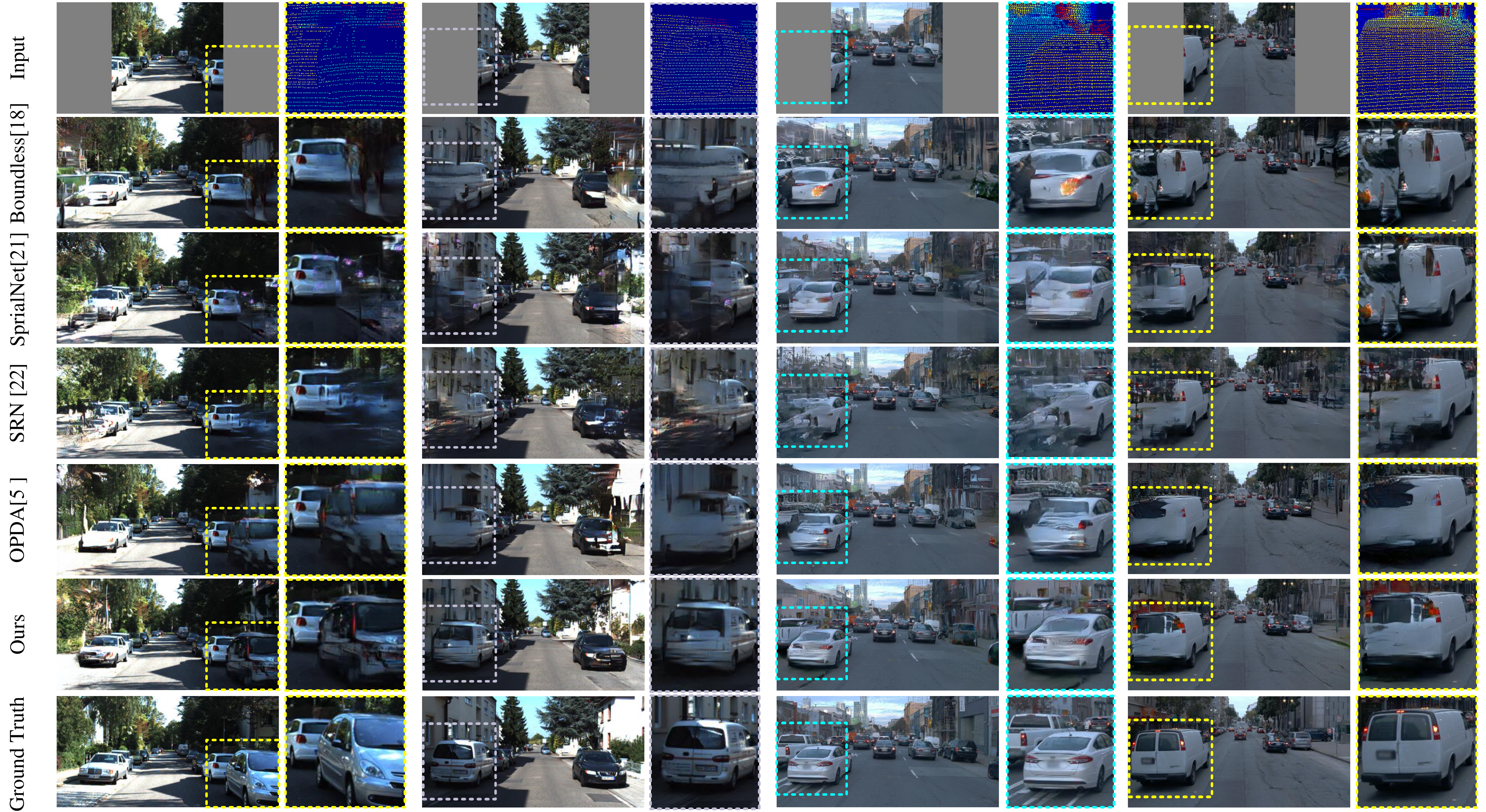}
  \centering  
  \setlength{\abovecaptionskip}{-0.2cm}  
  \caption{Qualitative results comparison with the state-of-the-art methods.}
  \label{experiment-1}
\end{figure*}

\setlength{\tabcolsep}{1.5mm}
\begin{table*}[h]\scriptsize

    \begin{minipage}{0.6\textwidth}
        \centering
        \renewcommand{\arraystretch}{1.0}
        \setlength{\abovecaptionskip}{-0.00cm} 
        \setlength{\belowcaptionskip}{-0.2cm}
        \caption{\scshape{Quantitative comparison on KITTI\cite{Uhrig2017THREEDV} and Waymo\cite{sun2020scalability}.}}
        \label{tab1-KITTI}
        \begin{tabular}{l|c c c c||c c c c}
            \hline\hline
         	Methods & \multicolumn{4}{c||}{KITTI} & \multicolumn{4}{c}{WAYMO} \\ \cline{2-9}
              & FID$ \downarrow $ &AP$ \uparrow $& PSNR$ \uparrow $& LPIPS$ \downarrow $ & FID$ \downarrow $ &AP$ \uparrow $& PSNR$ \uparrow $& LPIPS$ \downarrow $\\
            \hline\hline
            SpiralNet\cite{guo2020spiral} & 45.56 & 66.72\% & 15.02 & 0.2553 & 42.56 & 68.68\% & 22.62 & 0.2169\\
            SRN\cite{wang2019wide} & 34.99 & 68.13\% & 15.25 & 0.2477 & 42.85 &  69.84\% & 22.31  & 0.2149\\
            Boundless\cite{teterwak2019boundless} & 32.13 & 67.96\% & 15.13 & 0.2544 & 38.97 & 69.83\% & 20.68 & 0.2298\\ 
            OPDA\cite{zhang2021image} & 26.80 & 74.75\% & 15.51 & 0.2342 & 37.67 & 71.66 \% & 21.75 & 0.2203\\ 
            Ours & \bfseries 25.52 & \bfseries 79.23\% & \bfseries 16.27 & \bfseries 0.2182 & \bfseries 33.10 & \bfseries 74.03\% & \bfseries 22.67 & \bfseries 0.1964\\
            \hline
        \end{tabular}
	\end{minipage} 
	\hspace{.15in}
\begin{minipage}{0.30\textwidth}
	    \centering
	    \renewcommand{\arraystretch}{1.2}
	    \setlength{\abovecaptionskip}{-0.00cm} 
        \setlength{\belowcaptionskip}{-0.2cm}

        \caption{\scshape{Comparison in AP3D(R40).}}
        \label{tab-3d}
        \begin{tabular}{l|ccc}
            \hline
        	Methods & Bbox & BEV & 3Dbox\\
            \hline\hline
            SpiralNet\cite{guo2020spiral} & 96.40 & 94.25 & 94.20\\
            SRN\cite{wang2019wide} & 96.11 & 94.14 & 93.98\\
            Boundless\cite{teterwak2019boundless} & 93.66 & 95.38 & 93.51\\ 
            OPDA\cite{zhang2021image} & 96.46 & 96.24 & 94.13\\ 
            Ours &  \bfseries 96.53 &  \bfseries 96.40 & \bfseries 94.30\\
            
            \hline
        \end{tabular}
\end{minipage}
\end{table*}
\vspace{-5pt}
\section{Experiments}
\label{sec4}
\vspace{-5pt}
\subsection{Experimental Settings}
\label{s41}
\vspace{-5pt}
We train our DGONet on the street datasets: the KITTI dataset\cite{Uhrig2017THREEDV} and
the Waymo dataset\cite{sun2020scalability}.
And we set the known area size to 256 $ \times $ 256, and expand to 256 $ \times $ 512 in both sides. Our metrics contain three standard image evaluation, $i.e.$ the FID\cite{heusel2017gans}, LPIPS\cite{zhang2018unreasonable} and PSNR. 
To evaluate the realistic implications of our work as distinct from general outpainting methods, we introduce the evaluation metric: Average Precision(AP) from 2D object detection\cite{cai2018cascade} to reflect the credibility of the outpainting and restoration performance objectively. 
Likewise, we also consider multimodal 3D object detection\cite{chen2022focal} metric:AP3D(R40)\cite{simonelli2019disentangling}.
\begin{table}\scriptsize
    \centering
    \setlength{\abovecaptionskip}{0pt} 
    \setlength{\belowcaptionskip}{-0.2cm}
    \caption{\scshape{The contribution of each part on Waymo\cite{sun2020scalability}.}}
    \label{tab1-ablation-KITTI}
    \begin{tabular}{c c c c c|c c c}
        \hline
         & \textit{ML} &\textit{DGF} & ${{\mathcal{L}}_{edge}}$ & ${{\mathcal{L}}_{cm}}$ 
        & FID$ \downarrow $ &AP $ \uparrow $& LPIPS $ \downarrow $\\
        \hline\hline
        Baseline &  &  &  &  & 36.89 & 72.24\% & 0.2105\\
        Baseline &\scriptsize\Checkmark & \scriptsize\Checkmark  &\scriptsize\Checkmark & & 35.51 & 73.63\% & 0.2088\\ 
        Baseline & & \scriptsize\Checkmark & \scriptsize\scriptsize\Checkmark  & \scriptsize\Checkmark & 35.50 & 73.16\% & 0.2093\\ 
        Baseline &\scriptsize\Checkmark  &  & \scriptsize\Checkmark  & \scriptsize\Checkmark & 34.56 & 73.08\% & 0.1972\\ 
        Baseline &\scriptsize\Checkmark & \scriptsize\Checkmark & \scriptsize\Checkmark & \scriptsize\Checkmark & \bfseries 33.10 & \bfseries 74.03\% & \bfseries 0.1964\\
      \hline
    \end{tabular}
\end{table}
\vspace{-5pt}
\subsection{Quantitative Evaluations}
\vspace{-5pt}
Table \ref{tab1-KITTI} shows the results of our method and the state-of-the-art image outpainting methods\cite{teterwak2019boundless, guo2020spiral, wang2019wide, zhang2021image} in the KITTI and Waymo datasets. 
Our approach performs the best, especially on learning-based perception similarity measures FID and LPIPS, which indicate our images are more consistent with human perception and recognition. 
And the improvement of AP greatly exceeds the comparative approaches and verifies the reliability of our results. 
3D modality is a primary role in FSCN\cite{chen2022focal}, while RGB modality is used as an aid to provide features from 2D. So our higher image quality can be clearly demonstrated in the comprehensive AP3D (R40) score in Table \ref{tab-3d}.
Our approach provides reliable RGB for performance improvement in multimodal tasks.
The representation that the images we draw can be detected by computers also makes our approach more meaningful in practice.
\vspace{-10pt}
\subsection{Qualitative Comparisons}
\label{s42}
\vspace{-5pt}
We provide qualitative comparison results in Fig.\ref{experiment-1}. 
The general outpainting methods cannot reconstruct objects at seams in complex street scenes.
And even with the addition of depth, OPDA\cite{zhang2021image} cannot generate a plausible scene structure either. Instead, our approach accurately continues the style and content of the image at the boundary. It establishes appropriate structures based on the features learned from the cross-modal fusion.
From the enlarged parts, our approach structurally restores the spatial hierarchy represented by depth modalities, such as vehicles, trees, buildings, and ground.
\vspace{-5pt}
\subsection{Ablation Studies}
\label{s43}
\vspace{-5pt}
To validate the effectiveness of each component, we investigate ablation studies and show in Table \ref{tab1-ablation-KITTI}. 
When all components are utilized simultaneously, the model achieves the best results on all metrics.
\vspace{-5pt}
\section{Conclusions}
\label{sec5}
\vspace{-5pt}
In this paper, we explore a novel reliable image outpainting task combining depth modality and propose the Depth-Guided Outpainting Network.
We explicitly rely on the differences and unique characteristics of the two modalities to design the \textit{Multimodal Learning Module}. 
And the \textit{Depth Guidance Fusion Module} is proposed to realize a progressive cross-modal fusion scheme to facilitate the extrapolated establishment.
Furthermore, we design the additional \textit{Cross-modal Loss} and \textit{Edge Loss} to enhance the reliability of the results. 
The comprehensive experiments, 2D and 3D object detection, demonstrate that our method enables image outpainting with more realistic and believable performance.

\footnotesize
\balance
\bibliographystyle{IEEEbib}
\bibliography{sample}
\end{document}